\documentclass[11pt,a4paper]{article}
\usepackage{coling2020}
\usepackage{times}
\usepackage{latexsym}

\colingfinalcopy

\usepackage{url}
\usepackage[utf8x]{inputenc}
\usepackage[pdftex]{graphicx}
\usepackage[all]{xy}
\usepackage{amsmath,amssymb,multirow,color}
\usepackage{algorithmic}
\usepackage[algoruled,linesnumbered,noend,noline]{algorithm2e}
\usepackage{multirow}

\usepackage{etoolbox}
\apptocmd{\thebibliography}{\raggedright}{}{}

\usepackage{dashrule}

\usepackage{pict2e,tikz-cd}

\DeclareRobustCommand{\doublemultimapsymbol}{%
  \begingroup\setlength{\unitlength}{\fontcharht\font`A}%
  \roundcap
  \put(0.0,0.4172){\rotatebox{180}{\tiny $\multimap$}}
  \put(0.29,0.0){\tiny $\multimap$}
  \endgroup
  \;\;\;
}

\usepackage{rotating}
\usepackage{cleveref}

\title{Predicting Word Similarity in Context with Referential Translation Machines}

\author{Ergun Bi\c{c}ici \\
            AI Enablement \\
            Huawei Türkiye R\&D Center \\
            Istanbul, Turkey \\
              \url{orcid.org/0000-0002-2293-2031} \\
              {\tt \url{ergun.bicici@huawei.com}} \\
              {\tt \url{bicici.github.io}}
}

\date{}

\begin{document}

\maketitle


\begin{abstract}
We identify the similarity between two words in English by casting the task as machine translation performance prediction (MTPP) between the words given the context and 
the distance between their similarities. We use referential translation machines (RTMs), which allows a common representation for training and test sets and stacked machine learning models.
RTMs can achieve the top results in Graded Word Similarity in Context (GWSC) task.
\end{abstract}

\section{Grading the Similarity of Words within Context}

Graded Word Similarity in Context (GWSC) task~\cite{semeval2020_Task3_GCWS} is about the similarity of two words graded in continuous scale in two different contexts $c_1$ and 
$c_2$ defined by 
$60$ words on average and the change in the word pair similarity (wps) when the context is changed from $c_2$ to $c_1$. 
The subtasks are about unsupervised prediction of wps in both $c_1$ and $c_2$ and of the change in wps, which need not be found directly subtracting the two wpses. 
GWSC provides only the two words and the two contexts. 
The word pairs used are from SimLex999 dataset~\cite{Simlex999} in English, Croatian, Finnish, and Slovenian and we only participate in the task in English.

GWSC is a reverse engineering task in the sense that the actual wps in different contexts are being predicted as if read, scored, and evaluated by humans. 
SimLex999 scores are in the range $[0,10]$ where $10$ means the words are similar, $0$ dissimilar, and $5$ neutral.
Additionally, with unsupervised learning, the task involves optimization before for obtaining the target scores that mimic wps and approach SimLex999 type similarity scores and after for 
obtaining consistent scores that fit in the score range and distance or the change in wps, which is a subtask of the task.

\begin{figure}[t]
\centering
\includegraphics[width=1.0\linewidth]{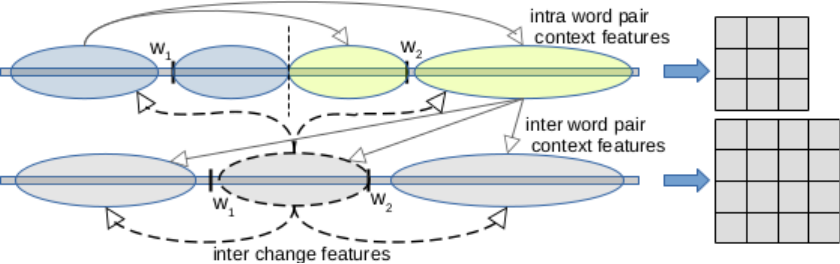}
\caption{Intra and inter context wps averages. Intra score averages pairs with words on different sides of the context and inter averages all 3x3.}
\label{WordSimilarityContextImage}
\end{figure}

We obtain features to measure the inter and intra similarity of the two contexts, $c_1$ and $c_2$. The intra similarity divides each context into 4 regions and builds an averaged similarity 
matrix using the semantic similarity databases to obtain the similarity of word pairs among words appearing in different regions. 
The inter similarity divides each context into 3 regions and again builds an averaged similarity matrix but computing for word pairs appearing in paired regions of the two contexts. 
Therefore, in the inter similarities, both the contexts and the words are paired. 
Intra word pair context is used for modeling the wps and inter word pair context is used for modeling the change in the similarity in two contexts. 
Both model semantic polarity or attraction and \Cref{WordSimilarityContextImage} depicts both.

Intra context wps (intra-cwps) features measure the positive or negative semantic attraction between the regions surrounding the word pair. 
In language modeling, a similarly named feature, statistical lexical attraction~\cite{Yuret98}, refers to the gain in the information to encode a sequence of words by bits using the mutual 
information of the probability that some paired words occur more likely in the same relative positions. 
Statistical lexical attraction is about the relatedness of words, which can be turned into a binary classification task, related or not, but GWSC is about wps, which can be turned into a ternary 
classificaion task: similar, dissimilar, and neutral. The averaged intra-cwps scores approach the SimLex999 type of similarity.

Inter context wps (inter-cwps) features measure the contextual changes in the semantics surrounding the target word pair in the two contexts and they are used to model the change in wps even if both 
contexts have the same score. As mentioned, semantic relatedness and similarity are different and word pairs are separated accordingly in the Wordsim-353 dataset~\cite{Wordsim353}. 
The wps datasets used~\cite{Simlex999,Wordsim353,Simverb3500,RW2013,Card660,MENDataset2014} and their size are in~\Cref{DataSize}. We refer to this combined dataset as WPSIMDAT.
By using a monotonicity assumption, we obtain a nonminimal transitive closure of WPSIMDAT, which expanded its size to 110K pairs. 
The monotonicity assumption transitively computes the wps of $w_i$\doublemultimapsymbol$w_k$ if transitivity conditions in ~\Cref{WPSIMDATClosure} hold.

\begin{table}[t]
\begin{center}
\begin{tabular}{@{\hspace{0.0cm}}c@{\hspace{0.1cm}}c@{\hspace{0.0cm}}}
source & \# of word pairs\\
\hline
Wordsim-353 & 203 \\
SimLex999  & 1000 \\
Simverb-3500 & 3500 \\
MEN         & 3000 \\
card-660    & 660 \\
rw          & 2034 \\
\hline
\end{tabular}
\end{center}
\caption{wps data sources and sizes (WPSIMDAT).}
\label{DataSize}
\end{table}

\begin{table}[t]
\begin{center}
{
\begin{tabular}{@{\hspace{0.0cm}}l@{\hspace{0.0cm}}}
if ($w_j$\doublemultimapsymbol$w_k) > w_i$\doublemultimapsymbol$w_j$ $\wedge$ $w_j$\doublemultimapsymbol$w_k$ $> 5$): \\
\hspace*{0.5cm} if $( w_i$\doublemultimapsymbol$w_j > 5 )$: \\ 
\hspace*{1.0cm} s $= w_i$\doublemultimapsymbol$w_j + (w_j$\doublemultimapsymbol$w_k - w_i$\doublemultimapsymbol$w_j) w_j$\doublemultimapsymbol$w_k / 10$ \\
\hspace*{1.0cm} $w_i$\doublemultimapsymbol$w_k = \min ( s, 10 )$ \\
\hspace*{0.5cm} if $( w_i$\doublemultimapsymbol$w_j < 5 \wedge w_j$\doublemultimapsymbol$w_k > 5 )$ \\
\hspace*{1.0cm} s $= w_i$\doublemultimapsymbol$w_j + (w_j$\doublemultimapsymbol$w_k - w_i$\doublemultimapsymbol$w_j) w_j$\doublemultimapsymbol$w_k / 10$ \\
\hspace*{1.0cm} if $s < 5$ then $w_i$\doublemultimapsymbol$w_k = \min ( s, 10 )$
\end{tabular}
}\end{center}
\caption{Transitivity conditions of WPSIMDAT.}
\label{WPSIMDATClosure}
\end{table}

Our processing use intra-cwps for modeling the target score and inter-cwps for modeling the change. As depicted in~\Cref{WordSimilarityContextImage}, intra-cwps score averages pairs with words on 
different sides of the context and inter-cwps averages all 3x3. GWSC is an unsupervised learning task where the target similarity scores are provided only for the 10 word pairs in the practice data. 
The target wps is obtained with the overall average of intra-cwps. 
The third feature set is the inter wps change features and they are used to obtain a score for the change with the difference between the averaged wps difference of intra and inter regions where the 
average of two in the first context is subtracted from the average of two in the second context. The target change in wps is obtained with its average over the 6 scores using ternary region splits.
For items without any word pairs found in WPSIMDAT, we use the output of a quadratic polynomial function, $a x^2 + b x + c$, that fits the intra-cwps scores. 
When intra-cwps scores are the same, we average the new score with the previous scores with weight $0.1$ since the contexts are different and therefore we expect to observe nonzero change. 


We also use wps semantic score using WordNet~\cite{WordNet} and NLTK~\cite{NLTK} provided Jiang-Conrath sense similarity measure using the information content from the WMT parallel corpora English 
side as the LM corpus. These WordNet similarity (WNsim) scores are mainly used as features to approximate wps score in both inter and intra models.
In the prediction stage, we predict both intra-cwps and inter-cwps scores. 

\subsection{wps Features}

We prepare 145 features for modeling semantic similarity (\Cref{wpsfeatures}). Edit distance is the Levenshtein distance between strings, length is measured in characters, and maximum or minimum is 
between all word pairs in the corresponding context regions. Since we have 3x3 regions in inter-cwps, together with their row and column averages, we obtain 16 features. We also obtain the wps values 
after mean / min / max of the regional mean / min / max values since a single word can affect the semantics of the whole region or context. 

\begin{table}[t]
\begin{center}
\begin{tabular}{@{\hspace{0.0cm}}c@{\hspace{0.1cm}}l@{\hspace{0.1cm}}c@{\hspace{0.0cm}}}
type & description & \# \\
\hline
inter & edit distance, length ratios for regions & 30 \\
inter & mean simlex for 3 regions and overall & 4 \\
inter & max of max WNsim & 16 \\
inter & mean of max WNsim & 16 \\
inter & mean of min WNsim & 16 \\
inter & min of min WNsim & 16 \\
inter & mean of mean WNsim & 16 \\
\hline
intra & edit distance, length ratios for regions & 20 \\
intra & mean of mean WNsim in $c_1$ & 1 \\
intra & mean of mean WNsim in $c_2$ & 1 \\
intra & ratio of mean WNsims in $c_1 / c_2$ & 1 \\
intra & mean of max WNsim in $c_1$ & 1 \\
intra & mean of max WNsim in $c_2$ & 1 \\
intra & ratio of max WNsims in $c_1 / c_2$ & 1 \\
intra & mean of min WNsim in $c_1$ & 1 \\
intra & mean of min WNsim in $c_2$ & 1 \\
intra & ratio of min WNsim in $c_1 / c_2$ & 1 \\
intra & multiplication of the three ratios & 1 \\
\hline
change & prediction by the quadratic function & 1 \\
\end{tabular}
\end{center}
\caption{There are 145 wps features.}
\label{wpsfeatures}
\end{table}

\subsection{Unsupervised Learning of Word Pair Similarity}

Due to the unsupervised nature of the task, we are not provided the labels for the word pairs, which are actually from the SimLex-999 dataset we used to derive the features.
Even though we did not use the wps scores directly from the dataset, their scores are indirectly included as a component averaged when calculating the intra and inter cwps scores. 
In the end, we find an artificial wps score that mimics the actual wps scores according to WPSIMDAT and to our contextual model of score averaging for wps and its difference in two different 
contexts. We refer to this score as awpss, artificial wps score. 
The scores of the training and test data are similar with mean, max, min on the training set are $5.7355$, $10$, $0.0686$ and on the test set are $5.3704$, $9.8$, $0.417$.
The change in all of the 360 word pairs including the practice set is on average $-0.55$ and $0.94$ in absolute terms in contrast, the change in the practice data is $1.159$ on average and $1.771$ 
after taking the absolute values of the change. The wps scores we obtain in the practice set is $5.725$, which are close where we achieve the best scores.
Even though SimLex-999 wps values are used within the awpss computations, they were also not included specifically for the word pair in question for the two contexts but we included all word pairs 
found from WPSIMDAT in the computation. The rankings show that RTM can achieve top results in GWSC.

\begin{table*}[t]
{
\small
\begin{center}
\begin{tabular}{@{\hspace{0.0cm}}c@{\hspace{0.1cm}}|c@{\hspace{0.1cm}}c@{\hspace
{0.1cm}}c@{\hspace{0.1cm}}c@{\hspace{0.1cm}}c@{\hspace{0.1cm}}c@{\hspace{0.1cm}}
c@{\hspace{0.1cm}}c@{\hspace{0.0cm}}}
English & prac. change & prac. score & eval. change & eval. score & post-eval. change & post-eval. score  \\
\hline
ranks & 1 & 3 & 13 & 9 & 4 & 5 \\
out of & 3 & 5 & 14 & 15 & 4 & 5\\
\hline
\end{tabular}
\end{center}
}
\caption{RTM ranks in GWSC at SemEval-2020.}
\label{TestsetRanksGWSC}
\end{table*}

\subsection{Predicting the Intensity of the Structure and Content in Tweets}

Affect in tweets~\footnote{\url{www.twitter.com}} task (Task 1)~\cite{semeval2018_Task1_affect_in_tweets} is about predicting the intensity of the emotion expressed for tweets within sadness, joy, fear, or anger emotion categorizations or the valance (sentiment). The emotion within tweets is about how the tweeter wrote and valence or sentiment is about what the tweeter wrote. 
Intensity scores are obtained with best-worst scaling (bws)~\cite{LREC18-TweetEmo}, which counts only the number of times a tweet is labeled as best or worst among 4 tweets where each is annotated by multiple workers. The scores are obtained with the percentage of counts scaled to $[-1, 1]$, which is later scaled to $[0, 1]$ for the task. bws can decrease the annotation effort to obtain the set of binary comparisons used to obtain reliably agreed labels.\footnote{bws is similar to inter-annotator agreement (IAA) $\tau$, which is in $[-1, 1]$ and calculated as IAA $\tau = (C - D) / (C+D)$, where C is the number of concordant pairs and D is the number of discordant pairs. IAA $\tau$ is the same measure as ``Kendall's $\tau$ with ties penalized'' used in~\cite{WMT2013} (Task1.2), where it was used for measuring the correlation between quality estimation systems and human rankings.
Oracle METEOR evaluation achieved $\tau=0.23$ for ranking in 2013's quality estimation task (Task 1.2, see \cite{WMT2013}). 
We tried a randomized IAA (RIAA) $\tau$~\cite{QTLeap} where the calculations are similar to IAA $\tau$, but all ties were converted into \emph{better} and \emph{worse} ranking randomly to obtain more robust results by distributing the ambiguous counts. This also helps normalization of the counts such that their sum are same for different models. A corresponding randomized bws would randomly assign the remaining pairwise comparison counts, where for 4 tweets only a single comparison would be left ambiguous, and the count can be distributed evenly.}

We model the task as MTPP of the tweets to the emotions to answer questions like ``to what degree is this tweet showing the emotion of''. Since a single emotion word need not provide enough context for semantic discrimination, we use sets of words for each emotion that express the same meaning using a subset of the WordNet affect emotion lists~\cite{Strapparava2004}. 
The lexicon used for English is in \Cref{wordnetlexicon}. We obtained their translations to Arabic and Spanish using web translation sites~\footnote{e.g. \url{translate.google.com} or \url{www.bing.com/translator}} to obtain the corresponding lexicon.
We use the whole set of words corresponding to the emotion instead of the emotion word to translate to.
For valence intensity tasks, we used both of the sets of words from emotions joy and sadness as a single sentence to translate to for the tweet's MTPP. 
Task 1 predicts the emotion and valence intensity in Arabic, English, and Spanish tweets where the evaluation metric is Pearson's correlation.~\footnote{The program for evaluation is at 
\url{https://github.com/felipebravom/SemEval_2018_Task_1_Eval}}

\subsection{Predicting the Attributes that Discriminate the Semantics}

Capturing discriminative attributes task (Task 10, SemEval-2018)~\cite{semeval2018_Task10_discriminative_attributes}
is looking at whether an attribute (e.g. red) can be used to discriminate between two other words (e.g. apple and banana) to complement semantic similarity efforts. The answer to whether the 
attribute can be used to discriminate the two words can be useful for semantic similarity with contextual dependency. 
The task is posed as a binary classification task: $f(w_1, w_2, a) \rightarrow 0 \; ? \; 1$ where $F_1$ is used for evaluation of target predictions that are either $0$ or $1$ showing whether the 
attribute can be used for discrimination for the given context. 
RTMs are used via casting the task as MTPP between the words and the attribute and building predictors that use the distance between the predictions. 
We assume that the discriminative power increase when the attribute is similar to words with significant difference.
We apply similar approach in GWSC where we use the correponding contexts instead of the attribute and add a row for each: $w_1 \rightarrow c_1$, $w_2 \rightarrow c_2$.

\section{Stacked RTM Models for Predicting the Discriminative Power of Attributes}

\begin{figure*}[t]
\centering
\includegraphics[width=1.0\linewidth]{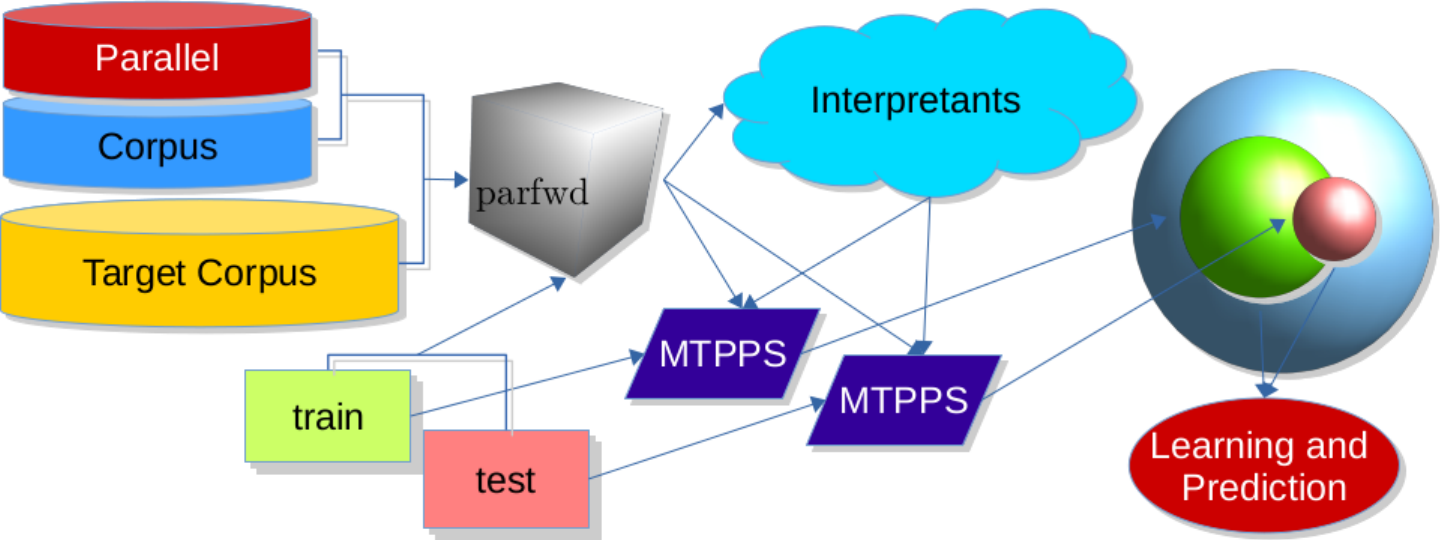}
\caption{RTM depiction: \texttt{parfda} selects interpretants close to the data using corpora; 
two MTPPS use interpretants, training data, and test data to generate features in the same space; 
learning and prediction use these features as input where spheres represent feature spaces.}
\label{RTMDiagram}
\end{figure*}

We use referential translation machine (RTM) models~\cite{Bicici:RTM_SEMEVAL} for predicting wps in GWSC, which use \texttt{parfda}~\cite{Bicici:ParFDA:WMT2016} to select both parallel and 
monolingual data close to the task instances selected specifically for the task, which are referred as interpretants, to derive features measuring the closeness of the test sentences to the training 
data, the difficulty of translating them, and to identify translation acts between any two data sets using machine translation performance prediction system (MTPPS)~\cite{Bicici:MTPP:MTJ2013,Bicici:MTPPS:SNCS2022} to 
build prediction models. Interpretants provide context and text for feature derivation to link translation source and target and training and test sets.
Interpretants are selected from the corpora distributed by the news translation task of WMT~\cite{WMT2017,WMT2019} and they consist of monolingual sentences used to build the LM and parallel sentence 
pair instances used by MTPPS to derive the features. We built RTM models using:
\begin{itemize}
\item 250 thousand sentences for training data
\item 5 million sentences for LM
\end{itemize}
RTMs are applicable in different domains and tasks and in both monolingual and bilingual settings. 
Figure~\ref{RTMDiagram} explains RTMs' model building process where machine learning models including 
ridge regression (RR), support vector regression (SVR), AdaBoost~\cite{AdaBoost}, and extremely randomized trees (TREE)~\cite{ExtrementRandomizedTrees} 
in combination with feature selection (FS)~\cite{GuyonWBV02} and partial least squares (PLS)~\cite{PLS1984} are used. 
We use averaging of scores from different models for robustness~\cite{Bicici:RTM:SEMEVAL2017}. 
Model implementations use \texttt{scikit-learn}.~\footnote{\url{http://scikit-learn.org/}}
We optimize $\lambda$ for RR, $\gamma$, C, and $\epsilon$ for SVR using grid search, minimum number of samples for leaf nodes and for splitting an internal node for TREE, the number of features for 
FS, and the number of dimensions for PLS. We use 500 estimators in the TREE model and also for AdaBoost. 
We evaluate with Pearson's correlation ($r$), mean absolute error (MAE), relative absolute error (RAE), MAER (mean absolute error relative), and 
MRAER (mean relative absolute error relative)~\cite{Bicici:RTM_SEMEVAL}.
We use $7$-fold cross-validation on the training set to rank models.

RTMs generate features for the training and the test set to map both to the same space where the total number of features in Task 1~\cite{semeval2018_Task1_affect_in_tweets} becomes 492 and Task 10 
becomes 117~\cite{semeval2018_Task10_discriminative_attributes}. The difference is due to the smaller context the attribute word provides and most of the sentence-level features become not useful 
including the sentence structure parsing features or word alignment features.


\begin{figure*}[t]
\centering
\includegraphics[width=0.9\linewidth]{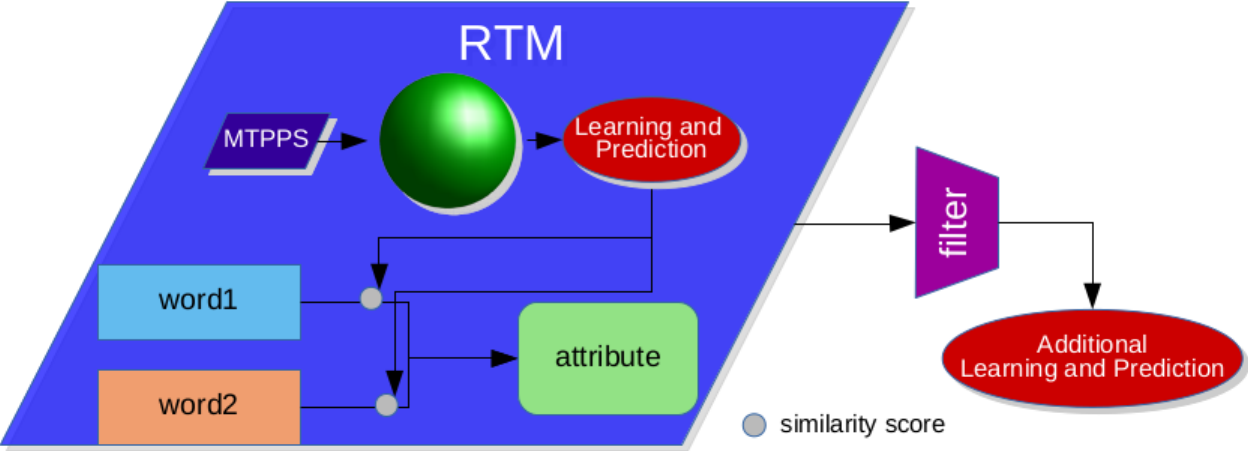}
\caption{RTM with stacked combined prediction use a combined model to obtain feature representations and predictions for $w_1 \rightarrow a$ and $w_2 \rightarrow a$, which are processed before 
additional learning and prediction.}
\label{RTM_combined_predictor}
\end{figure*}

The stacked RTM model with combined prediction step (\Cref{RTM_combined_predictor}) use the same model to predict the MTPP similarity of the translations of $w_1$ or $w_2$ to 
$a$, $w_1 \rightarrow a$ and of $w_2 \rightarrow a$, where the data is collected such that the first row for each attribute is for $w_1 \rightarrow a$ and the second is for 
$w_2 \rightarrow a$ where $a$ provides the context. Stacking is used to build higher level models using predictions from base prediction models where they can also use the 
probability associated with the predictions~\cite{Stacking1999}. The combined model in \Cref{RTM_combined_predictor} is adding the predictions as additional feature.
We obtain an RTM representation vector for each of the instances by using the derived features and the combination of the prediction scores along with 5 additional features:
\begin{align}
\hat{y}_1 & & \hat{y}_2 & & |\hat{y}_1 - \hat{y}_2| & & (\hat{y}_1 + \hat{y}_2) / 2 & & \sqrt{\hat{y}_1*\hat{y}_2}
\end{align}
After this filtering step, we run another learning and prediction on the concatenation of the features from both rows and the additional features.

\begin{table}[t]
\begin{tabular}{l}
\hspace*{5cm}$1$\&$2$-gram wrec \\
\hspace*{5cm}$1$\&$2$\&$3$-gram wGM \\
\hspace*{5cm}$1$\&$2$-gram wGM \\
\hspace*{5cm}$1$\&$2$-gram w$F_1$ \\
\hspace*{5cm}$1$-gram wGM
\end{tabular}
\caption{Top $5$ features selected for Task 10}
\label{top5features_Task10}
\end{table}

\begin{table}[t]
\begin{tabular}{l}
\hspace*{5cm}translation logprobability bpw \\
\hspace*{5cm}word alignment (1 - WER) \\
\hspace*{5cm}word alignment $F_1$ score \\
\hspace*{5cm}$3$gram w$F_1$ \\
\hspace*{5cm}sentence number of characters \\
\end{tabular}
\caption{Top $5$ features selected for Task 1 Spanish}
\label{top5features_Task1}
\end{table}

RTM results in the Task 1~\footnote{\url{https://competitions.codalab.org/competitions/17751}} and Task 10~\footnote{\url{https://competitions.codalab.org/competitions/17326}} competitions are in 
\Cref{TestResultsNew} and ranks are in \Cref{TestsetRanks}~\cite{semeval2018_Task1_affect_in_tweets}. 
8 of the results obtain MRAER larger than 1 suggesting more work towards these tasks or subtasks. weight \# models combine top \# models' predictions~\cite{Bicici:RTM:SEMEVAL2017}.
The predictions for Task 10 were transformed to binary classes by thresholding with $0.5$ and obtains $0.47$ $F_1$. 

The top $5$ features selected for Task 10 are listed in \Cref{top5features_Task10}.
$1$\&$2$gram w$F_1$ is $F_1$ score over $1$-gram and $2$-gram features with recall computed according to the sum of the likelihood of 
observing them among $1$-grams or $2$-grams correspondingly (wrec) and precision computed according to all corresponding counts in $n$-grams. wGM is weighted geometric mean of the arguments of $F_1$. The features enable linking $w_1$ and $a$ and $w_2$ and $a$ and $8$ of the top $10$ features use $n$-gram features, which makes sense for linking words and we observe this even semantically for Task 10.
The top $5$ features selected for Task 1 are listed in \Cref{top5features_Task1}. bpw is bits per word. WER is word error rate.

\begin{table*}[t]
{
\small
\begin{center}
\begin{tabular}{@{\hspace{0.0cm}}c@{\hspace{0.1cm}}|c@{\hspace{0.1cm}}c@{\hspace
{0.1cm}}c@{\hspace{0.1cm}}c@{\hspace{0.1cm}}c@{\hspace{0.1cm}}c@{\hspace{0.1cm}}
c@{\hspace{0.1cm}}c@{\hspace{0.0cm}}}
& T1 EI-en & T1 EI-ar & T1 EI-es & T1 V-en & T1 V-ar & T1 V-es & T10 \\
\hline
ranks & 44 & 13 & 10 & 35 & 13 & 10 & 9 \\
out of & 50 & 15 & 17 & 39 & 15 & 15 & 9 \\
\hline
\end{tabular}
\end{center}
}
\caption{RTM ranks at SemEval-2018.}
\label{TestsetRanks}
\end{table*}

\begin{table*}[t]
{
\begin{center}
\begin{tabular}{@{\hspace{0.0cm}}llll|lllll@{\hspace{0.0cm}}}
\multicolumn{3}{c}{Task} & & $r$ & MAE & RAE & MAER & MRAER \\ 
\hline
English & Task 1 & emotion & anger & 0.245 & 0.1689 & 1.086 & 0.3619 & 1.074 \\ 
English & Task 1 & emotion & fear & 0.05 & 0.1526 & 1.044 & 0.413 & 0.952 \\ 
English & Task 1 & emotion & joy & 0.028 & 0.1641 & 1.053 & 0.438 & 0.963 \\ 
English & Task 1 & emotion & sadness & 0.004 & 0.1566 & 1.043 & 0.4064 & 0.944 \\ 
English & Task 1 & emotion & ALL & 0.2245 & 0.1734 & 1.14 & 0.367 & 1.1693 \\ 
Arabic & Task 1 & emotion & anger & 0.209 & 0.1413 & 0.992 & 0.3093 & 0.878 \\ 
Arabic & Task 1 & emotion & fear & 0.173 & 0.1444 & 1.01 & 0.3112 & 0.905 \\ 
Arabic & Task 1 & emotion & joy & 0.377 & 0.1417 & 0.94 & 0.4126 & 0.805 \\ 
Arabic & Task 1 & emotion & sadness & 0.269 & 0.1442 & 0.983 & 0.3291 & 0.872 \\ 
Arabic & Task 1 & emotion & ALL & 0.2543 & 0.1428 & 0.9771 & 0.344 & 0.8545 \\ 
Spanish & Task 1 & emotion & anger & 0.183 & 0.1706 & 1.005 & 0.3807 & 0.927 \\ 
Spanish & Task 1 & emotion & fear & 0.398 & 0.1607 & 0.92 & 0.4548 & 0.846 \\ 
Spanish & Task 1 & emotion & joy & 0.298 & 0.1676 & 0.945 & 0.4856 & 0.843 \\ 
Spanish & Task 1 & emotion & sadness & 0.405 & 0.1513 & 0.909 & 0.3943 & 0.838 \\ 
Spanish & Task 1 & emotion & ALL & 0.324 & 0.1627 & 0.9426 & 0.4311 & 0.8568 \\ 
\hline
English & Task 1 & valence & & 0.1326 & 0.1884 & 1.0399 & 0.525 & 0.9791 \\ 
Arabic & Task 1 & valence & & 0.2981 & 0.1879 & 0.9366 & 0.5637 & 0.8482 \\ 
Spanish & Task 1 & valence & & 0.2152 & 0.1684 & 0.973 & 0.5399 & 0.8317 \\ 
\hline
\end{tabular}
\end{center}
}
\caption{RTM results on the test set.}
\label{TestResultsNew}
\end{table*}

\Cref{TestResultsNew} lists the latest results obtained after the challenge. 
The number of results with MRAER larger than 1 decreased to 2 from 8.
The MRAER obtained by RTMs in STS in 2016 is $0.73$~\cite{Bicici:RTM:SEMEVAL2016} and in quality estimation task for English to German in 2017 is $0.76$~\cite{Bicici:RTM:WMT2019}.
The predictions for Task 10 were transformed to binary classes by thresholding with optimized thresholds on the training set. 

\section{Conclusion}


Referential translation machines obtain automatic prediction of semantic similarity using MTPP. 
We presented encouraging results with stacked RTM models for GWSC with our novel MTPP modeling for translation to context, for predicting the 
intensity of the structure and content in text with MTPP modeling for translation to WordNet emotion lists, and for the discriminative power of attributes using stacked RTM models.
Our results also enable comparisons of prediction results of RTMs in different natural language processing tasks. 
%
%

%

%

\bibliography{ebiciciPubs,Bibtex}

\pagebreak

\appendix

\section{English Lexicon used from WordNet Affect Emotion Lists}
\label{wordnetlexicon}

English lexicon used for identifying discriminative attributes (Task 10) as a subset of the WordNet affect emotion lists~\cite{Strapparava2004} available at \url{http://web.eecs.umich.edu/~mihalcea/affectivetext/}.
~\footnote{The original lexicon was made available freely for research purposes. The lexicon in this section is for the publication and demonstration.}

{\small
\hspace*{-0.5cm}
\begin{tabular}{@{\hspace{0.0cm}}c@{\hspace{0.0cm}}}
\textit{anger} \\
\parbox{8.75cm}{torment pissed frustrated crucify pout grudge umbrageous incensed spitefulness execration exacerbate choleric infuriating displease teasing aggression misology maliciously riled ire baffled spiteful pestered enragement rag misanthropy misogynism harried grudging annoyance abominate displeased furious irritating wrothful nettlesome detest vexation hatred indignantly angry offense vindictiveness despising revengefully disdain belligerent murderousness stung misanthropical lividly madness irritation belligerently misogynic begrudge abhor temper indignation infuriated anger discouraged grievance contemn bitterness vexatious rancour resentment hate exasperating malign covetous enraged envious enviously jealous irascibility loathing displeasingly gravel odium grasping infuriate sulkiness outrage annoyed mad covetously vindictively belligerence prehensile enviousness scene harassment murderously enfuriate wrathful hateful displeasure loathe scorn maliciousness execrate venom enviably pique roiled heartburning malicious tantrum irritated envy frustration nettle harassed spleen brood enviable offend animosity despiteful maddening sore balked aggressive enmity vexed vex jealously resentfully angered galling pestering vindictive hackles umbrage bothersome score gall misanthropic vexing persecute aggressiveness indignant amuck oppress rancor jealousy annoying resentful covetousness frustrating outraged tantalize devil greedy hostilely wrathfully covet exasperation frustrate hostility misogyny incense rage choler malevolently malignity malevolence vengefulness huffiness malevolent furiously chafe pesky begrudging huffy angrily harass wroth despise fury rile avaricious malefic bother abomination spite nettled antagonism aggravated hatefully aggravate hostile belligerency annoy maddened provoked exasperate malice plaguey vengefully displeasing lividity stew pestiferous evil wrath irritate infuriation grizzle aggravation abhorrence misoneism fit} \\
\textit{fear} \\
\parbox{8.75cm}{atrocious anxiously diffidently coldheartedness cruelty frightening hesitantly chill dreadfully scarily panic foreboding hideous panicky hesitancy timidness hesitance cower intimidation ruthlessness dreaded alarm suspense intimidate chilling dreadful horrify diffidence alert fearful presentiment creep apprehension hysterical uneasily anxious bashfully horribly timid apprehensive ugly hardheartedness cringe unkind browbeaten horrifying fawn shadow unassertiveness horrifyingly dread awfully intimidated frightful afraid apprehensiveness outrageous premonition panicked shy scary frighten hysterically heartlessly pitilessness shyly frighteningly fearfulness diffident monstrously dismay timorous fear horrified timorousness unassertive fright cruelly scared consternation affright apprehensively dire timidity heartlessness heartless timidly scarey hysteria alarmed horrific cowed presage shyness mercilessness awful trepidation hideously unsure scare bullied frightened horridly fearsome crawl creeps direful terrified fearfully suspensive shuddery} \\
\end{tabular}
}

{\small
\hspace*{0.35cm}
\begin{tabular}{@{\hspace{0.0cm}}cc@{\hspace{0.0cm}}}
\textit{sadness} \\
\parbox{8.75cm}{disconsolate sorrow guilt dolorous demoralization weight dispirit remorse oppression heartbreaking sorrowfully drear grievously ruefulness sorry oppressed despondently despondency wretched woefulness dreary desolation bereft downcast attrition forlornly sadness plaintive pitiable woebegone penitently pitying desolate heartsickness demoralize dispiritedness dysphoria depressive loneliness persecuted remorsefully sadly dark joylessly regretful despairingly tyrannical suffering sadden ruthfulness distressed melancholic sad bored discouraged disheartening regret penitentially plaintively heartache forlorn tearfulness lachrymose ruefully cheerlessness forlornness gloomily oppressive contrite shamefaced heartrending drab plaintiveness melancholy gloomful mournfulness penance brokenheartedness saddening repentant dysphoric depress mourning heartsick repentantly heavyhearted doleful aggrieve contritely hapless rue contrition downheartedness depression tyrannous piteous pitiful uncheerful demoralized laden tearful grievous woeful gloomy penitence godforsaken repent penitent rueful pathetic miserably dolefully bad oppressiveness dismay repentance cheerless dingy heavyheartedness joyless grim grief persecute bereaved grieving grieve disconsolateness lamentably unhappiness guilty oppress helplessness weeping misery oppressively deject hangdog heartbreak demoralising woefully depressed uncheerfulness contriteness dispirited joylessness disheartened poor gloominess depressing remorseful down unhappy low weepiness mournful deplorably blue glum despondent harass gloom downhearted dolefulness mournfully demoralizing woe downtrodden compunction shamed sorrowing dolourous misfortunate dismal sorrowful glooming sorrowfulness cheerlessly dispiriting} \\
\textit{joy} \\
\parbox{8.75cm}{kick identification amicably anticipation cheerfully favourably eager close gratifying suspenseful triumphant captivation beaming protectiveness devoted zest ardour satisfy exhilarating occupy preference exciting happily penchant compatibly jubilancy tenderness near exultingly approved satisfactory merry becharm crush admire joy exult jolly admiration adoration complacent pleased eagerly anticipate fondness regard kindhearted proudly happiness entrance rapport enamoredness affectional keenness closeness praiseworthily hilariously festive affect appreciated exultant elation exhilaration protective prideful gloat impress liking unworried pride friendly soothe move thrill gratify enthusiastically catch gleefully joyous excitement teased titillate gleefulness joyousness uproarious goodwill gaiety great worship worry glad favorably comfortably taste exhilarate perkiness uplift empathy satisfying hearten attachment comfort gratifyingly beguile commendable fulfil affection tender fulfillment titillation sunny jocularity stimulating warmth concern cheer enthralling joyfully comfortableness entranced approving mirthfully content comforting jubilantly predilection jocund console euphoria satisfactorily merrily satisfied partiality strike kindly gladfulness enjoy solace good fascination love zealous sympathetically benevolent fulfilment affectionateness devotedness gladdened riotously buoyancy benevolently enthralled adorably beneficent hilarity amorous warmhearted likable weakness belonging carefree avidness carefreeness affective captivate barrack recreate sympathetic protectively euphoriant urge friendliness jubilant comfortable benefic intimacy admirably enthusiasm respect sympathy esteem bang emotive enchant approval charm favour satisfyingly amative fascinating gloating fancy brotherlike hilarious loyalty lovingly capture bewitching satisfiable beneficed compatibility lovesome identify beneficially zeal cheery revel favourable jubilance fulfill happy lovingness gayly brotherhood jollity congratulate complacence satisfaction lightsomeness romantic intoxicate joyously triumph charitable devotion heart ebullient benevolence exhilarated exuberance approve contented captivated exuberant euphoric exultantly festal rush cheerful endearingly tickle jubilation complacency gladsomeness enthusiastic delighted triumphal joyful affectionate gloatingly charmed eagerness empathic amicability amatory flush kid charge screaming rejoice relish entrancing fondly expansively exuberantly contentment inspire fond like suspensive likeable amicable triumphantly expectancy}
\end{tabular}
}

\end{document}